\documentclass[11pt]{article}

\usepackage[preprint]{acl}

\usepackage{times}
\usepackage{latexsym}

\usepackage[T1]{fontenc}

\usepackage[utf8]{inputenc}

\usepackage{microtype}

\usepackage{inconsolata}

\usepackage{graphicx}

\usepackage{adjustbox}
\usepackage{diagbox}
\usepackage{slashbox}
\usepackage{multirow}
\usepackage{booktabs}
\usepackage{caption}
\usepackage{subcaption}
\usepackage{makecell}
\usepackage{enumitem}
\setlist{nosep,leftmargin=2.5em}
\usepackage{wrapfig}
\usepackage{float}
\usepackage{tikz}
\usetikzlibrary{calc}
\usepackage{courier}
\usepackage{color}
\usepackage{xcolor}
\usepackage{colortbl}
\usepackage{pifont}
\definecolor{customblue}{rgb}{0.1, 0.1, 0.8}
\definecolor{tablegreen}{RGB}{46, 160, 100}
\hypersetup{
    colorlinks=true,
    citecolor=customblue,
    linkcolor=customblue,
    urlcolor=customblue
}

\usepackage[most]{tcolorbox}
\tcbuselibrary{listings}
\tcbuselibrary{breakable}
\tcbuselibrary{skins}
\tcbset{mybreakable/.style={
    breakable,
    enhanced,
}}
\tcbset{promptbox/.style={enhanced, colback=white, colframe=#1!75!black,
    coltitle=white, fonttitle=\small\bfseries, top=2pt, bottom=2pt, left=2pt, right=2pt}}
\tcbset{promptboxtop/.style={promptbox=#1, sharp corners=south, after skip=0pt}}
\tcbset{promptboxbottom/.style={promptbox=#1, sharp corners=north, before skip=-0.4pt,
    toprule=0mm, colback=white, colbacktitle=#1!75!black,
    fonttitle=\small\bfseries}}
\newlength{\promptbreakindent}
\AtBeginDocument{\settowidth{\promptbreakindent}{\scriptsize\ttfamily-~}}
\lstdefinestyle{pythonstyle}{language=Python, frame=single,
    basicstyle=\scriptsize\ttfamily, columns=fullflexible, keepspaces=true,
    breaklines=true, breakindent=1.5em,
    linewidth=\dimexpr\linewidth-3.4pt\relax, xleftmargin=1.7pt, xrightmargin=0pt,
    keywordstyle=\bfseries\color[HTML]{008000}, stringstyle=\color[HTML]{BA2121},
    commentstyle=\itshape\color[HTML]{3D7B7B}, showstringspaces=false,
    emph={tour_meeting,cli}, emphstyle=\bfseries\color{customblue},
    literate={¥}{{\textyen}}1}
\lstdefinestyle{promptstyle}{basicstyle=\scriptsize\ttfamily, breaklines=true,
    columns=fullflexible, keepspaces=true,
    breakindent=\promptbreakindent, breakautoindent=true,
    moredelim=[s][\color{tablegreen!75!black}\bfseries]{\{}{\}}}
\usepackage{rotating}
\usepackage{cuted}

\usepackage{algorithmic}
\usepackage[linesnumbered, ruled, vlined]{algorithm2e}
\usepackage[linesnumbered, ruled, vlined]{algorithm2e}
\usepackage{etoolbox}
\makeatletter
\patchcmd\algocf@Vline{\vrule}{\vrule \kern-0.4pt}{}{}
\patchcmd\algocf@Vsline{\vrule}{\vrule \kern-0.4pt}{}{}
\patchcmd\@makecaption{\scshape}{}{}{}
\patchcmd\@makecaption{\\}{:~}{}{}
\patchcmd\@makecaption{\raggedleft}{}{}{}
\makeatother
\SetAlgoNlRelativeSize{-1}
\title{
\includegraphics[width=10cm]{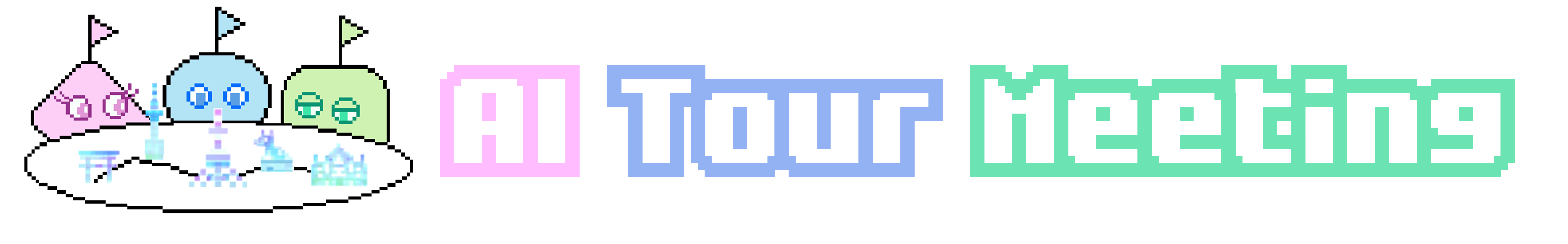}\\
AI Tour Meeting: Group Travel Planning by LLM Agents
}

\author{
Daisuke Kikuta \\
NTT, Inc. \\
\texttt{daisuke.kikuta@ntt.com} \\
}

\begin{document}
\maketitle
\begin{strip}
    \begin{minipage}{\textwidth}\centering
        \vspace{-2\intextsep}
        \setlength{\tabcolsep}{0.3em}
        \begin{tabular}{rl}
            \makebox[0.4cm][c]{%
              \raisebox{-0.2\height}{%
                \includegraphics[width=0.5cm]{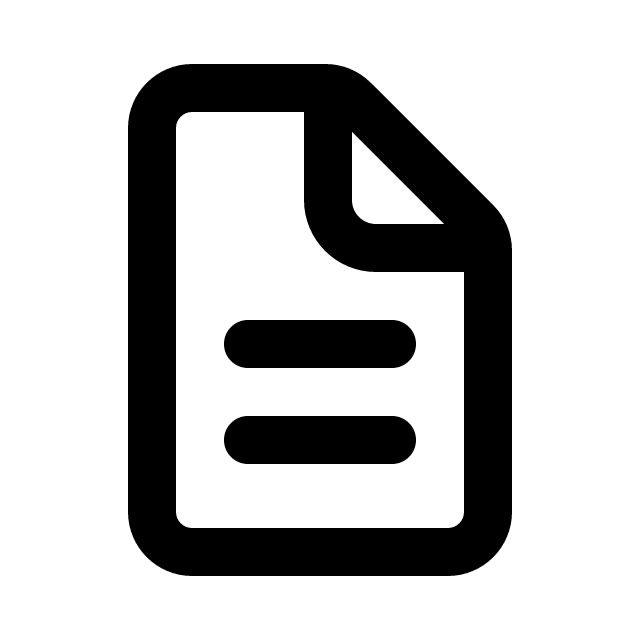}%
              }%
            } & \textbf{Docs} : \url{https://ntt-dkiku.github.io/ai-tour-meeting} \\
            \makebox[0.4cm][c]{%
              \raisebox{-0.15\height}{%
                \includegraphics[width=0.4cm]{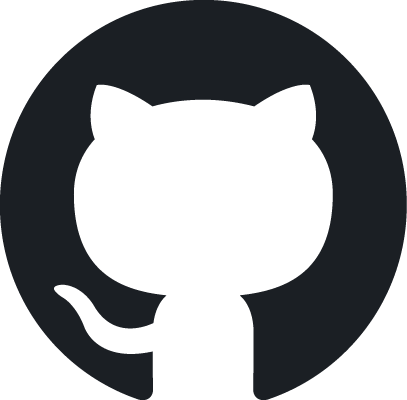}%
              }%
            } & \textbf{Code} : \url{https://github.com/ntt-dkiku/ai-tour-meeting} \\
            \makebox[0.4cm][c]{%
              \raisebox{-0.15\height}{%
                \includegraphics[width=0.4cm]{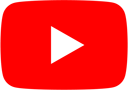}%
              }%
            } & \textbf{Video} : \url{https://x.gd/VS9RD} \\
        \end{tabular}
    \end{minipage}
\end{strip}
\begin{abstract}
This paper proposes AI Tour Meeting, a group travel planning framework powered by multiple Large Language Model (LLM)-based agents.
The agents are instantiated with distinct personas and collaboratively seek an itinerary that satisfies their constraints and preferences through natural language discussion.
The framework enables easy and flexible orchestration of such discussions by providing interfaces for configuring agent personas, discussion workflows, monitoring, and LLM deployment. Its primary use case is a simulation tool for analyzing the behavior of multiple LLM agents during tour planning discussions.  
This paper demonstrates the utility of the framework by presenting system validation and several analytical results obtained by the framework.
\end{abstract}

\section{Introduction}
Group travel planning aims to find an itinerary that balances multiple objectives among multiple participants, such as cost, travel time, and individual preferences.
Prior work has focused on supporting human groups in determining balanced itineraries~\cite{NguyenRicci2018,AMT-TRE,grouptravelbench}.
Meanwhile, in broader domains, recent studies have explored the use of persona-based Large Language Model (LLM) agents to simulate human behavior~\cite{generative-agents,sotopia,personallm,hu-collier-2024-quantifying}.
This emerging direction is also expected to be relevant to group travel planning.
For example, automatic evaluation in such simulation environments reduces the costs of large-scale behavioral analysis in group travel planning and the evaluation of existing group travel recommender systems.
In application, users could interact with LLM agents representing group members who are unable to participate in the actual discussion and collaboratively determine an itinerary with them.

To address this emerging need, we propose AI Tour Meeting, a group travel planning framework powered by multiple LLM agents. 
Each agent is instantiated with a distinct persona and collaborates with the others to find an itinerary that satisfies their constraints and preferences through natural-language discussion.
Our framework provides interfaces to orchestrate this process.
Compared with existing persona-based agent simulations~\cite{generative-agents,sotopia} and general-purpose multi-agent frameworks~\cite{li2023camel,wu2024autogen}, our framework provides domain-specific abstractions for group travel planning, including a structured itinerary representation, explicit proposal-voting workflows, and constraint validation and metric monitoring specialized for travel planning.
This allows users to focus on their use cases without implementing the required functionality from scratch using a general-purpose framework.

In the following, we describe the technical details of our framework and validate whether it can successfully complete discussions using Qwen 3.5 and GPT models of different sizes.
We then demonstrate its effectiveness as a simulation tool through analysis examples conducted using the framework.

\begin{figure*}
  \centering
  \includegraphics[width=\textwidth]{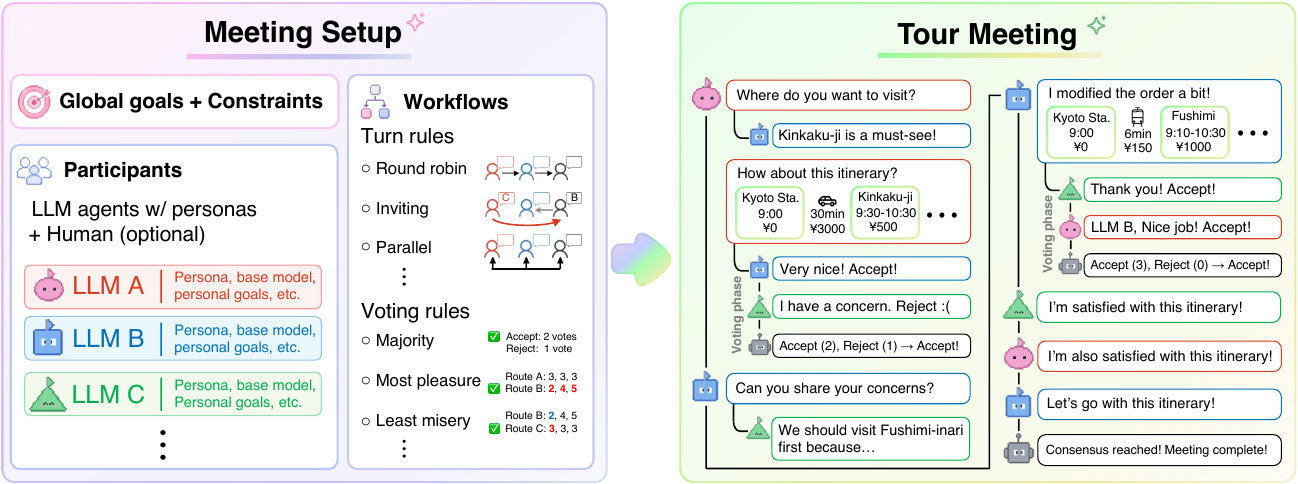}
  \caption{An overview of AI Tour Meeting. The user first configures the meeting settings, including the global goals, constraints, participants, and workflow, and then starts the meeting. The meeting alternates between conversation and voting phases whenever a new itinerary is proposed. Once all participants are satisfied with the currently accepted itinerary, the meeting concludes.
}
  \label{fig:teaser}
\end{figure*}

\section{Proposed Framework: AI Tour Meeting}
Figure \ref{fig:teaser} shows the overview of our framework: we first set up both LLM participants with distinct personas and a discussion workflow, then start the meeting among them aligned with the specified workflow.
The meeting begins with the conversation phase, in which participants take turns searching information, asking questions, and proposing itineraries.
Once an itinerary is proposed, it transitions to the voting phase, where the participants decide whether to accept or reject it.
This two-phase process continues until all participants are satisfied with the currently accepted itinerary.

In this section, we describe the definitions of itineraries and participants, the workflow options, the monitored metrics, and the user interface.

\subsection{Itinerary}
An itinerary proposed during the meeting consists of information about destinations and the transportation between them, and is defined as follows:
\begin{equation}
    R = (d_1, d_2, \dots, d_{N}),
\end{equation}

where $R$ is the itinerary, $d_i$ is the $i$-th destination.
Each destination includes a name, description, monetary cost, arrival time, duration of stay, transportation mode from the previous destination, transportation cost, and transportation duration.

\subsection{Participant}
\label{sec:participant}
Each LLM agent (participant) is instantiated with a distinct persona. The persona consists of the following attributes: 
\textit{Name} specifies their name;
\textit{Background} specifies their experience or situation;
\textit{Personality} specifies their stable traits;
\textit{Preference} specifies their preferences related to the tour; 
\textit{Goal} specifies their personal goals in the tour;
\textit{Role} selects their role in the meeting from either facilitator or attendee;
\textit{Tone} specifies their speaking tone such as formal or friendly; 
\textit{Explanation style} selects a strategy for convincing others to accept their proposal, from either subjective explanation, contrastive explanation \cite{Kikuta_2024} or both. The former provides subjective opinions of the proposal based on their preferences. 
The latter explains the objective pros and cons by comparing the proposed route with the current route in terms of objective metrics such as cost, time, and number of destinations.
We can select an LLM that instantiates the participant with model parameters such as temperature and seed. 
Both commercial LLMs and local LLMs (via Ollama or vLLM) are supported,
Context management methods, including auto-compaction and history truncation, can be also configured individually for each participant.

\paragraph{Actions}
The $\mathrm{persona}$ is given to the LLM as a system prompt, and the LLM participant takes actions during the meeting while role-playing according to the given $\mathrm{persona}$.
The meeting proceeds in a turn-based manner, and each participant may take multiple actions during their single turn.
The action at each step $s$ within a turn is defined as follows:
\begin{equation}
    (a^{\mathrm{type}}_s, \mathrm{msg}_s, a_s) = p_m(\mathrm{persona}_m, H_{s-1}),
\label{eq:action}
\end{equation}
where $p_m$ is the $m$-th LLM participant, $a_s^\mathrm{type}$ is the action type, $\mathrm{mgs}_s$ is the message associated with the action, $a_s$ is the taken action, and $H_{s-1}$ is the meeting history, which includes system messages for managing the discussion workflow $\mathrm{msg}_\mathrm{sys}$ and the past actions taken by all the participants $((a_1^\mathrm{type}, \mathrm{msg}_1, a_1), \dots, (a_{s-1}^\mathrm{type}, \mathrm{msg}_{s-1}, a_{s-1}))$.
The action $a_s$ varies depending on the selected action type $a^\mathrm{type}_s$, as follows:

\definecolor{midcolor}{RGB}{230, 243, 255}    
\definecolor{convcolor}{RGB}{255, 240, 230}   
\definecolor{votecolor}{RGB}{230, 255, 235}   
\newcommand{\splitbg}[4][0.5]{%
\begin{tikzpicture}[baseline=(T.base)]
  \node[inner sep=3pt, outer sep=0pt] (T) {#4};

  \coordinate (B) at ($(T.south west)!#1!(T.south east)$);
  \coordinate (U) at ($(T.north west)!#1!(T.north east)$);

  \fill[#2] (B) rectangle (T.north east);
  \fill[#3] (T.south west) rectangle (U);

  \node[inner sep=1pt, outer sep=0pt] at (T.center) {#4};
\end{tikzpicture}%
}
\begin{equation}
    a_s = \left\{
    \begin{array}{ll}
        \rowcolor{midcolor}
        \mathrm{Search}(q_s)          & \text{if } a^{\mathrm{type}}_s = \mathrm{search} \\
        \rowcolor{midcolor}
        \mathrm{Ask}(p_{m'}, q_s)     & \text{if } a^{\mathrm{type}}_s = \mathrm{ask} \\
        \rowcolor{midcolor}
        \mathrm{Reflect}              & \text{if } a^{\mathrm{type}}_s = \mathrm{reflect} \\[4pt]
        \rowcolor{convcolor}
        \mathrm{Propose}(R_s)         & \text{if } a^{\mathrm{type}}_s = \mathrm{propose} \\
        \rowcolor{convcolor}
        \phi                           & \text{otherwise} \\[4pt]
        \rowcolor{votecolor}
        \mathrm{Score}(v_s)           & \text{if } a^{\mathrm{type}}_s = \mathrm{scoring} \\
        \rowcolor{votecolor}
        \phi                           & \text{otherwise.}
    \end{array}
    \right.
\end{equation}

There are two types of actions: intermediate and terminal actions. Intermediate actions may be taken multiple times within a turn, whereas once a terminal action is taken, the current turn ends.
The \colorbox{midcolor}{intermediate} action types include $\mathrm{search}$, $\mathrm{ask}$, and $\mathrm{reflect}$.
$\mathrm{search}$ issues a query $q_s$ to retrieve information such as location or cost of destinations via web search.
$\mathrm{ask}$ asks another participant $p_{m'}$ a question $q_s$ and receives a response.
$\mathrm{reflect}$ organizes and reviews the discussion conducted so far.
The \splitbg{votecolor}{convcolor}{terminal} action types differ between the conversation phase and the voting phase.
In the \colorbox{convcolor}{conversation} phase, they include $\mathrm{propose}$, $\mathrm{satisfied}$, and $\mathrm{pass}$.
$\mathrm{propose}$ proposes a new itinerary $R_s$.
$\mathrm{satisfied}$ indicates that $p_m$ is satisfied with the current route.
$\mathrm{pass}$ ends their turn without expressing a specific position.
In the \colorbox{votecolor}{voting} phase, they include $\mathrm{accept}$, $\mathrm{reject}$, and $\mathrm{score}$.
$\mathrm{accept}$ votes for the proposed itinerary, $\mathrm{reject}$ does not votes for the proposed itinerary, and $\mathrm{score}$ assigns a score $v_s$ to the proposed route.

\subsection{Discussion workflow}
A tour meeting consists of a global goal, participants, constraints, a turn rule, and a voting rule.
The global goal is a high-level objective shared among all participants, e.g., planning a one-day sightseeing tour in Tokyo.
Each participant advances the discussion in pursuit of their own personal goal while remaining aligned with the global goal.
Constraints restrict the itinerary in terms of travel date, budget, and time window.

The tour meeting proceeds according to the specified turn rule, under which participants sequentially take the actions defined in Eq. (4) multiple times within their turns.
This normal state is referred to as the conversation phase.
Once a participant proposes a itinerary (i.e., terminates the turn with $a^\mathrm{type}=\mathrm{propose}$), the process transitions to the voting phase. In this phase, all participants except the proposer cast votes on the proposed itinerary. After collecting all votes, the acceptance or rejection of the proposed route is determined in accordance with the specified voting rule. If the itinerary is accepted, the current itinerary is replaced with the proposed one; if it is rejected, the current itinerary remains unchanged.
The process then returns to the conversation phase, and the turns resume with the next participant after the proposer.

When all participants consecutively indicate satisfaction with the current route (i.e., terminate their turns with $a^\mathrm{type}=\mathrm{satisfied}$), consensus is considered reached, and the meeting concludes.
In practice, the termination condition of the meeting may also be constrained by a maximum number of turns or a maximum execution time, in addition to the unanimous agreement.

Algorithm \ref{alg:meeting_worklfow} organizes the discussion workflow described above. 
In the following, we describe turn rules and voting rules available in our framework.

\SetInd{0.5em}{0.78em}

\begin{algorithm}[!t]
    \caption{Tour meeting workflow}
    \label{alg:meeting_worklfow}
    \small
    \SetKwInOut{Input}{Input}
    \SetKwInOut{Output}{Output}
    \Input{Participants $\mathcal{P}=\{p_1, p_2, \dots, p_M\}$;\\
           Turn rule $\pi_{\mathrm{turn}}$; 
           Voting rule $\pi_{\mathrm{vote}}$
    }
    \Output{The final tour itinerary $R_{\mathrm{final}}$}
    \SetNlSty{textbf}{}{:}
    
    $R_{\mathrm{final}} \leftarrow \phi$; 
    $H \leftarrow \phi$; 
    $C \leftarrow 0$\\

    \While{$C < M$}{
        $p \leftarrow \pi_{\mathrm{turn}}(\mathcal{P})$\\
        \While{$a^{\mathrm{type}} \in \{\mathrm{search},\mathrm{ask}, \mathrm{reflect}\}$}{
            $(a^{\mathrm{type}}, \mathrm{msg}, a) 
            \leftarrow p(\mathrm{persona}, H)$\\
            $H \leftarrow H + (a^{\mathrm{type}}, \mathrm{msg}, a)$\\
            \If{$a^{\mathrm{type}}=\mathrm{propose}$}{
                    $R'\leftarrow a$
                }
        }
        \uIf{$a^{\mathrm{type}}=\mathrm{propose}$}{
            $\mathcal{V} \leftarrow \phi$; $C \leftarrow 0$\\
            \For{$\_ \leftarrow 1$ \KwTo $M-1$}{
                $p' \leftarrow \pi_{\mathrm{turn}}(\mathcal{P}\setminus p)$\\    
                \While{$a^{\mathrm{type}} \in \{\mathrm{search},\mathrm{ask}, \mathrm{refect}\}$}{
                    $(a^{\mathrm{type}}, \mathrm{msg}, a) 
                    \leftarrow p'(\mathrm{persona}, H)$\\
                    $H \leftarrow H + (a^{\mathrm{type}}, \mathrm{msg}, a)$\\
                }
                
                \If{$a^{\mathrm{type}}\in \{\mathrm{accept}, \mathrm{reject}\}$}{
                    $\mathcal{V}\leftarrow\mathcal{V}+a^{\mathrm{type}}$
                }
            }

            \If{$\pi_{\mathrm{vote}}(\mathcal{V})=\mathrm{accept}$}{
                $R_{\mathrm{final}} \leftarrow R'$\\
            }
        }
        \uElseIf{$a^{\mathrm{type}}=\mathrm{satisfied}$}{
            $C \leftarrow C + 1$
        }
        \Else{
            $C \leftarrow 0$
        }
    }
    \Return $R_{\mathrm{final}}$
\end{algorithm}

\begin{figure*}
  \centering
  \includegraphics[width=\textwidth]{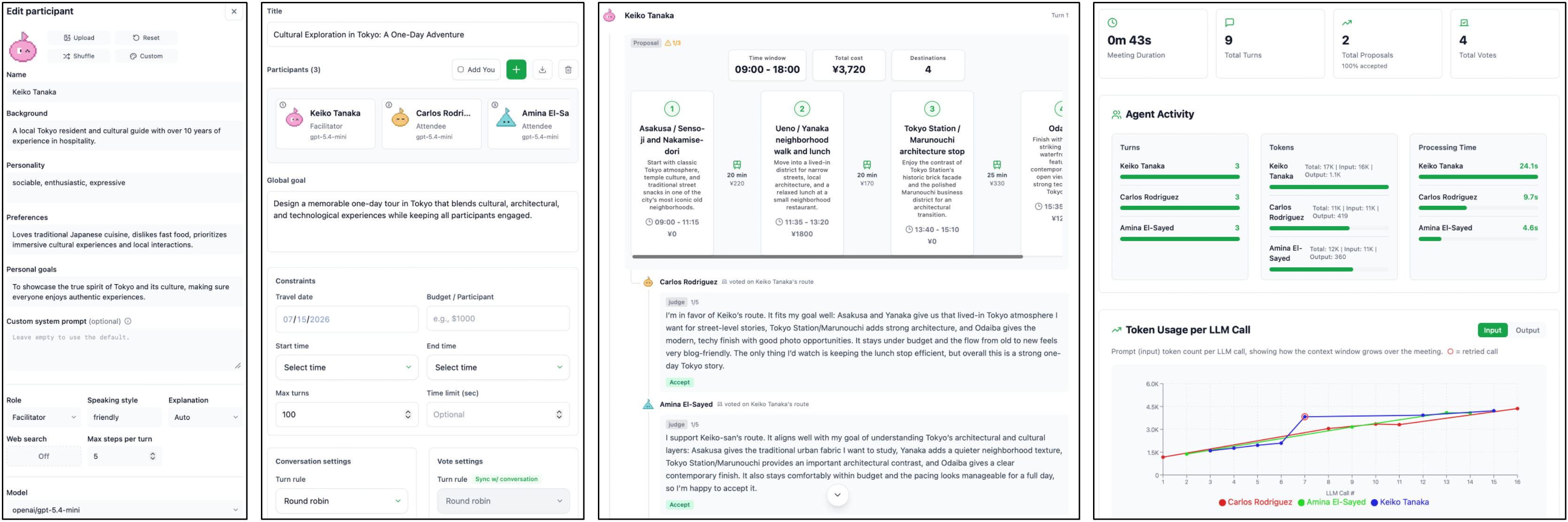}
  \caption{The graphical user interface of AI Tour Meeting. From left to right: the participant settings, the meeting settings, the ongoing meeting screen, and the analytics dashboard.}
  \label{fig:gui}
\end{figure*}

\subsubsection{Turn rule}
Our framework supports standard four turn rules, along with two additional options to make them more flexible.
There are five types of turn rules:
\textit{Round robin} rotates participants in a fixed cyclic order;
\textit{Inviting} lets the current speaker determine the next speaker;
\textit{Facilitating} assigns an additional turn to a facilitator after each turn, who then determines the next speaker;
\textit{Random} rotates participants randomly;
\textit{Parallel} lets all participants vote simultaneously (only for the voting phase).
Note that different turn rules may be applied separately to the conversation and voting phases.

In addition, when the \textit{Balancing} option is enabled, speaking turns are organized into cycles such that all participants take turns once per cycle to ensure fairness.
When the \textit{Volunteer} option is enabled, participants are allowed to select the $\mathrm{pass}$ as a terminal action and skip their turn (i.e., only those who wish to speak take a turn).

\subsubsection{Voting rule}
Our framework also supports standard five voting rules.
In \textit{Majority} and \textit{Unanimous}, participants cast a binary vote (accept or reject) on the proposed itinerary. The proposed itinerary replaces the current itinerary if it receives a majority of accept votes in majority voting, or if all participants vote accept in unanimous voting.
In \textit{Single decider}, a specified participant alone cast the binary vote, and the current itinerary is updated accordingly.
In \textit{Most pleasure} and \textit{Least misery}, participants assign a score to the proposed itinerary according to 10-point rubrics.
The proposed itinerary replaces the current itinerary if, in the former, the sum of all participants’ scores is greater than or equal to that of the current itinerary, or, in the latter, the minimum score across participants is greater than or equal to that of the current itinerary.

\begin{figure}[!t]
\begin{lstlisting}[style=pythonstyle]
from tour_meeting.cli import build_meeting

meeting = build_meeting(
  title="One-Day Tokyo Tour",
  global_goals="Plan a fun one-day Tokyo tour.",
  participants=[
    {
      "name": "Alice",
      "background": "A history enthusiast who ...",
      "goals": "Visit Senso-ji and ...",
      "model_name": "vllm/0/Qwen/Qwen3.5-9B",
      ...
    },
    ...
  ],
  constraints={"budget": "¥10K", "start": ...},
  settings={"max_turns": 100, "turn_rule": ...},
)

asyncio.run(meeting.run_cli())
\end{lstlisting}
\caption{Minimal example of Python API}
\label{fig:code}
\end{figure}

\begin{table*}[tb]
    \begin{center}
    \small
    \renewcommand{\arraystretch}{1.1}
    \begin{tabular}{l cccccccc}
        \toprule
        Model
        & \makecell{Completion\\ (\%) $\uparrow$}
        & \makecell{Consensus\\ (\%)}
        & \makecell{Constraint err. \\ (\%) $\downarrow$}
        & \makecell{Action fail. \\ (\%) $\downarrow$}
        & \makecell{Turns}
        & \makecell{Duration\footnotemark\\ (min)}
        & \makecell{Tokens [K] \\(in / out)}
        & Usable \\
        \midrule
        Qwen3.5-2B
        & $82$ & $58$ & $14.2$ & $1.3$
        & $61.5$
        & $72.5$
        & $7100$ / $114$
        & \textcolor{red!70!black}{\ding{55}} \\
        Qwen3.5-4B
        & $100$ & $100$ & $3.1$ & $0.5$
        & $15.7$
        & $37.0$
        & $240$ / $44$
        & \textcolor{green!55!black}{\ding{51}} \\
        Qwen3.5-9B
        & $100$ & $100$ & $0.0$ & $0.1$
        & $12.8$
        & $33.6$
        & $134$ / $37$
        & \textcolor{green!55!black}{\ding{51}} \\
        gpt-oss-20b
        & $100$ & $100$ & $0.4$ & $0.2$
        & $20.4$
        & $11.8$
        & $315$ / $25$
        & \textcolor{green!55!black}{\ding{51}} \\
        gpt-5.4-mini
        & $100$ & $80$ & $0.0$ & $0.1$
        & $34.5$
        & $6.3$
        & $535$ / $40$
        & \textcolor{green!55!black}{\ding{51}} \\
        \bottomrule
    \end{tabular}
    \end{center}
    \caption{System validation results across different models and model sizes. These results suggest that our framework requires an LLM with performance comparable to or higher than Qwen3.5-4B.}
    \label{tab:validation}
\end{table*}

\subsection{Monitoring and metrics}
Our framework collects various metrics to analyze both system-level performance and the dynamics of discussions.
The system-level metrics include agents’ token usage, the number of retries, and the number of time violations at destinations in the proposed itineraries.
The discussion-dynamics metrics include the number of meeting steps and turns, the message history, the distribution of executed action types, the distribution of votes and scores, and the details of the proposed itineraries.
Users can leverage these metrics to analyze discussions flexibly from multiple perspectives. Basic analyses, such as computing representative statistics and distributions for each metric, can be performed using built-in functions.

\subsection{User interface}
Figure \ref{fig:gui} shows the snapshots of meeting settings and meeting dialogue, and analytics dashboard in the GUI of AI Tour Meeting.
Participants and meetings can be easily configured through pull-down selections and text inputs, and the meeting can be browsed in a chatbot-style dialogue view. 
The analytics dashboard allows users to visually explore some basic analytical metrics.
Our framework also supports Python API.
As shown in Figure \ref{fig:code}, users can flexibly configure and launch meetings at scale with only a few lines of code.

\section{System Validation}
First, we validate whether our framework runs robustly across different models and model sizes.
\paragraph{Setup}
We use gpt-5.4-mini~\cite{gpt-5.4-mini} to generate 50 synthetic meetings with three participants whose preferences partially overlap and contain moderate conflicts that can be resolved through adjustment.
We then run these meetings using five different LLMs to instantiate the participants: Qwen3.5-2B/4B/9B~\cite{qwen3.5}, gpt-oss-20b~\cite{openai2025gptoss120bgptoss20bmodel}, gpt-5.4-mini (medium effort). 
The temperature is set to 0.6 for the local models and 1.0 for gpt-5.4-mini. Local models are served with vLLM~\cite{vllm} on a single RTX A6000 GPU (48GB).
All meetings are configured with round-robin turns, majority voting, a 100-turn limit, a time window of 09:00–18:00, and a budget of \$100 per participant.

\footnotetext{Since meeting duration largely depends on the machine used to deploy the LLMs, these values are for reference only.}

\paragraph{Results}
Table~\ref{tab:validation} shows the validation results. 
We report the following metrics:
\textit{completion} denotes the proportion of meetings that ended upon reaching consensus or the maximum number of turns, with at least one itinerary accepted;
\textit{consensus} denotes the proportion of meetings that ended upon reaching consensus;
\textit{constraint error} denotes the proportion of newly proposed itineraries that violated the constraints;
\textit{action failure} denotes the proportion of actions that failed after more than three retries; and
\textit{turns}, \textit{duration}, and \textit{tokens} denote the corresponding average values.
The results show that all models except for Qwen3.5-2B achieve a 100\% completion rate and the rates of constraint errors and action failures decrease as model performance improves.
Qwen3.5-2B produces several invalid actions, including repeating the same message and rejecting proposals by referring to a nonexistent “current route.”
Overall, our framework runs properly with LLMs that perform comparably to or better than Qwen3.5-4B.

\begin{table}[tb]
    \begin{center}
    \small
    \renewcommand{\arraystretch}{1.1}
    \begin{tabular}{l rrr}
        \toprule
        & Aligned & Mixed & Conflicting \\
        \midrule
        Turns & \cellcolor{tablegreen!5}$10.3_{\pm7.60}$ & \cellcolor{tablegreen!9}$12.1_{\pm9.80}$ & \cellcolor{tablegreen!40}$25.9_{\pm22.7}$ \\
        Proposals & \cellcolor{tablegreen!5}$2.3_{\pm2.30}$ & \cellcolor{tablegreen!8}$2.7_{\pm2.50}$ & \cellcolor{tablegreen!40}$6.5_{\pm5.30}$ \\
        Consensus rate & \cellcolor{tablegreen!40}$100\%$ & \cellcolor{tablegreen!40}$100\%$ & \cellcolor{tablegreen!5}$94\%$ \\
        Satisfaction & \cellcolor{tablegreen!40}$8.93_{\pm1.47}$ & \cellcolor{tablegreen!30}$8.39_{\pm1.70}$ & \cellcolor{tablegreen!5}$7.13_{\pm2.06}$ \\
        Victim rate & \cellcolor{tablegreen!40}$0.7\%$ & \cellcolor{tablegreen!38}$1.3\%$ & \cellcolor{tablegreen!5}$11.3\%$ \\
        \bottomrule
    \end{tabular}
    \end{center}
    \caption{Meeting results under controlled preference conflict among LLM participants.}
    \label{tab:alignment}
\end{table}

\begin{figure}[tb]
  \centering
  \begin{subfigure}[t]{\columnwidth}
    \centering
    \includegraphics[width=\linewidth]{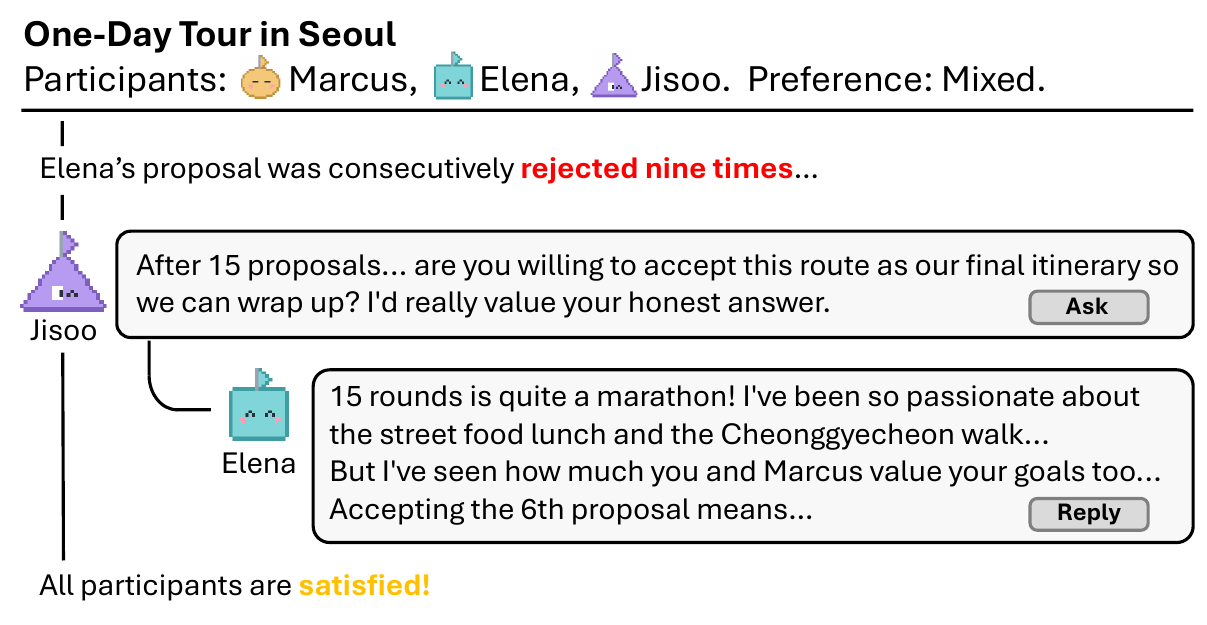}
    \caption{A pattern of prompting compromise.}
    \label{fig:case_seoul}
  \end{subfigure}\\[6pt]
  \begin{subfigure}[t]{\columnwidth}
    \centering
    \includegraphics[width=\linewidth]{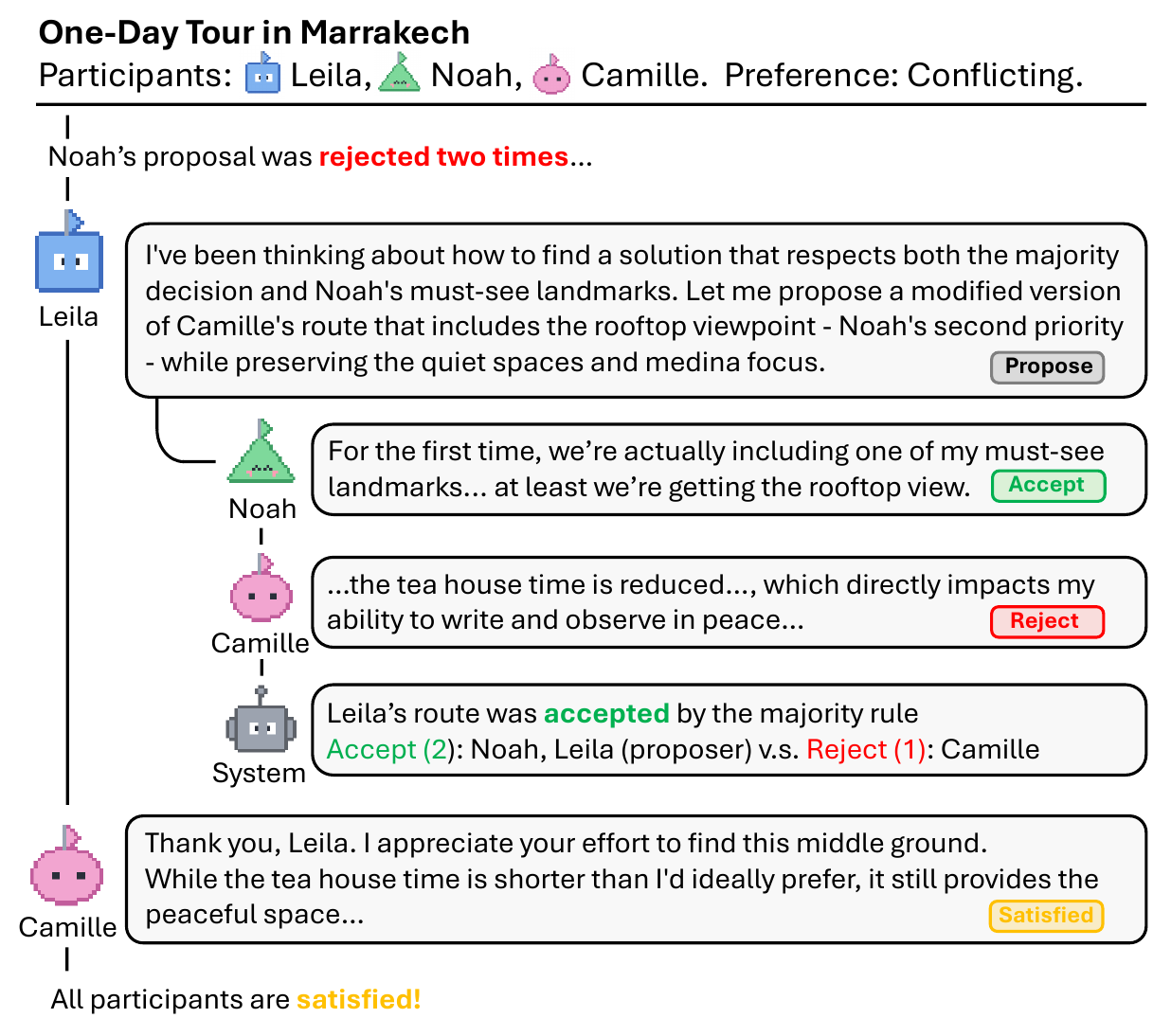}
    \caption{A pattern of mediation by another participant.}
    \label{fig:case_marrakech}
  \end{subfigure}
  \caption{Examples of consensus-building patterns.}
  \label{fig:case_studies}
\end{figure}

\section{Analysis Examples}
In this section, we demonstrate the utility of our framework by providing two analysis examples conducted using our framework.
\subsection{Preference conflict}
\label{sec:eval}
\paragraph{Setup}
This example analyzes how meeting dynamics change depending on the degree of preference conflict among participants.
We generate 50 synthetic meetings for each of three settings:
\textit{aligned} assigns aligned preferences among the three participants;
\textit{mixed} assigns mixed preferences as in the system validation;
\textit{conflicting} assigns conflicting preferences that require at least one participant to compromise on at least one aspect.
After each meeting, we also evaluate each participant’s satisfaction with the final itinerary on a 1–10 scale using an LLM-as-a-judge.
All meetings use the inviting turn rule and Qwen-3.5-9B as participants, with all other settings and constraints unchanged from the system validation.

\paragraph{Results}
Table~\ref{tab:alignment} shows that as the degree of preference conflict increases, the number of turns and itinerary proposals both increase, resulting in longer discussions. 
Meanwhile, as preference conflict increases, the consensus rate and average satisfaction score decrease, while the \textit{victim rate}---the proportion of participants with a satisfaction score of 4 or lower---increases.
These results suggest that discussions among LLM participants reproduce a natural phenomenon observed in human deliberation: greater conflict leads to longer discussions and makes it harder to find a proposal that satisfies everyone.

Figure~\ref{fig:case_studies} shows two representative patterns of the consensus-building process under preference conflict. 
Figure~\ref{fig:case_seoul} shows a case in which Elena’s proposals are repeatedly rejected by the other two participants. Eventually, Jisoo urges Elena to conclude the meeting, and Elena compromises to end the discussion.
Elena assigns the final itinerary a score of 3 in the post-hoc evaluation, indicating that she agreed despite remaining dissatisfied.
In contrast, Figure~\ref{fig:case_marrakech} shows a case in which, after Noah’s proposal was rejected twice, Leila intervened and proposed an itinerary on his behalf that incorporated his preferences.

\subsection{Speaking order}

\begin{figure}[tb]
  \centering
  \includegraphics[width=\columnwidth]{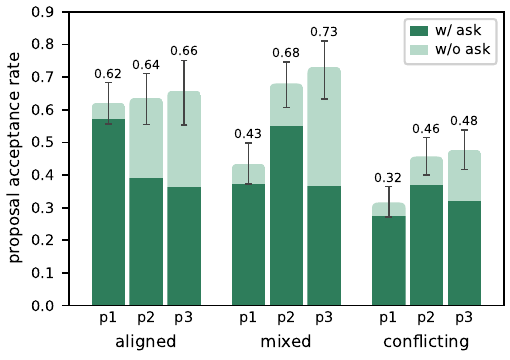}
  \caption{Proposal acceptance rates by preference conflict level and speaking position. Error bars are Wilson 95\% confidence intervals on the total rate. 
}
  \label{fig:turn_position}
\end{figure}

\paragraph{Setup}
This example analyzes how speaking order affects the proposal acceptance rate.
We run each of the 50 meetings at three different conflict levels under a round-robin turn rule with a Latin-square seat rotation: each meeting is executed three times, rotating the participants through the speaking positions $p_1$--$p_3$. The other settings and constraints are the same as the previous analysis example.

\paragraph{Results}
Figure~\ref{fig:turn_position} shows no statistically significant difference across speaking positions in \textit{aligned}. 
In \textit{mixed} and \textit{conflicting}, however, proposals from the second and third participants have significantly higher acceptance rates than those from the first participant.
This may be because participants have no preference conflict in \textit{aligned}, so proposals are likely to be accepted regardless of speaking order.
On the other hand, in \textit{mixed} and \textit{conflicting}, preference conflicts make it difficult for the first participant to propose an itinerary that satisfies the preferences of the others. Later participants can use more feedback from earlier proposals and voting results to propose itineraries that better reflect the preferences of others, resulting in higher acceptance rates.
Additionally, we found that most accepted proposals from the first participant were made after asking the others for input. In contrast, participants who spoke later could often make accepted proposals without asking, relying instead on information from the preceding discussion.

\section{Conclusion}
In this paper, we proposed a new LLM-based group travel planning framework and demonstrated its validity and utility through system-level evaluation and use cases.
We believe that our framework supports research on the behavior of LLM agents and automatic evaluation of recommender systems for group travel planning.
Beyond these uses, it may also enable new applications, such as recommender systems where LLM agents act on behalf of group members who are unable to participate and provide ideas from their perspectives (see Appendix~\ref{sec:interacitve}).


\bibliography{custom}

\clearpage
\begin{table*}[tb]
    \begin{center}
    \small
    \renewcommand{\arraystretch}{1.1}
    \begin{tabular}{l cccccccc}
        \toprule
        \# Participants
        & \makecell{Completion\\ (\%) $\uparrow$}
        & \makecell{Consensus\\ (\%)}
        & \makecell{Constraint err. \\ (\%) $\downarrow$}
        & \makecell{Action fail. \\ (\%) $\downarrow$}
        & \makecell{Turns}
        & \makecell{Duration\\ (min)}
        & \makecell{Tokens [K] \\(in / out)}
        & Validity \\
        \midrule
        $M=3$
        & $100$ & $100$ & $0.0$ & $0.1$
        & $12.8$
        & $33.6$
        & $134$ / $37$
        & \textcolor{green!55!black}{\ding{51}} \\
        $M=5$
        & $100$ & $100$ & $1.6$ & $0.8$
        & $25.4$
        & $149.3$
        & $659$ / $99$
        & \textcolor{green!55!black}{\ding{51}} \\
        $M=10$
        & $100$ & $96$ & $2.4$ & $0.3$
        & $68.6$
        & $471.5$
        & $3074$ / $406$
        & \textcolor{green!55!black}{\ding{51}} \\
        \bottomrule
    \end{tabular}
    \end{center}
    \caption{Additional system validation with Qwen3.5-9B across participant sizes $M \in \{3, 5, 10\}$}
    \label{tab:validation_size}
\end{table*}
\appendix

\section{License}
Our code is licensed under NTT's proprietary license, which restricts the use of our code to research purposes only. 
Since our framework is designed primarily as a simulation tool to support research on group travel planning, this restriction does not pose a significant limitation.

\section{Related Work}
\paragraph{Group travel planning}
\citet{NguyenRicci2018} proposed a system that supports consensus building by monitoring chat discussions and adapting recommendations to evolving user preferences.
\citet{AMT-TRE} proposed a method that integrates member relationships and POI information through attention mechanisms to optimize group itinerary recommendation.
More recent work has investigated whether LLMs can help group travel planning while resolving preference conflicts among group members~\cite{grouptravelbench}.
These works focus on supporting humans, whereas our work explores the use of LLM agents as proxies for humans in group travel planning.

\paragraph{Travel planning with LLM agents}
Recent work has proposed benchmarks for evaluating the travel-planning capabilities of LLMs~\cite{travelplanner,singh-etal-2024-personal}, as well as methods for improving their performance~\cite{vaiagemultiagentsolutionpersonalized,choi2026atlas}. 
However, these studies primarily focus on individual travel planning.

\paragraph{LLM-based multi-agent simulation}
LLM-based multi-agent simulation has been widely studied~\cite{generative-agents,sotopia,personallm,hu-collier-2024-quantifying}. 
However, to our knowledge, no prior work has specifically addressed such discussion in group travel planning.
From an implementation perspective, although general-purpose multi-agent frameworks~\cite{li2023camel,wu2024autogen} provide basic functionality for orchestrating agent interactions, users must implement domain-specific components for group travel planning. Our framework provides these components by default, allowing users to focus on their use cases.

\section{Human Participant (Interactive Mode)}
\label{sec:interacitve}
\begin{figure}[tb]
  \centering
  \includegraphics[width=\columnwidth]{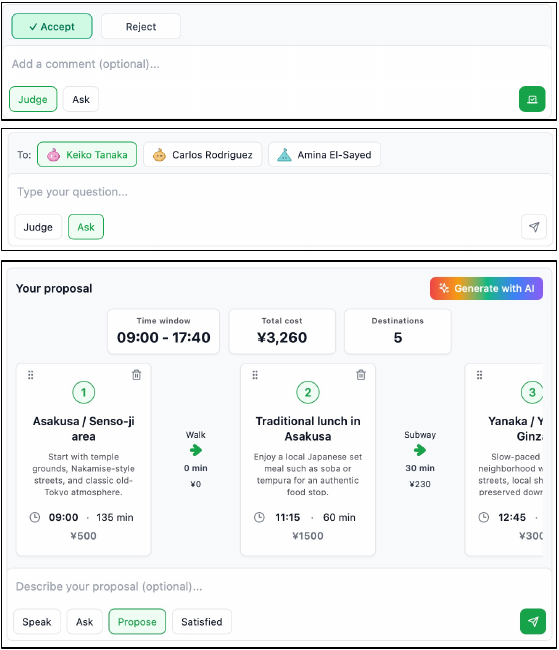}
  \caption{Chat boxes for human participants.}
  \label{fig:interactive}
\end{figure}

AI Tour Meeting also allows human users to participate in meetings. At their turn, they can interact with the LLM participants through the chat box by executing the same actions defined in Eq. (\ref{eq:action}) as the LLM participants. Figure~\ref{fig:interactive} shows snapshots of the chat box during voting, when asking another participant a question, and when proposing an itinerary. When proposing an itinerary, users can manually modify destinations. By clicking the “Generate with AI” button, they can interact with an LLM and have it refine their proposal.
A use case of this feature is to represent friends who are unable to attend the meeting as LLM participants based on their personas. By discussing the itinerary with these agents, users can incorporate the absent participants’ perspectives into the tour planning.

\section{Additional Validation: Participant Size}
\label{sec:additional_validation}
\paragraph{Setup}
We additionally validate whether the framework correctly completes meetings as the number of participants increases. 
We generate 50 synthetic meetings for each participant size (5 and 10 participants) using GPT-5.4-mini, and evaluate them with Qwen3.5-9B under the same settings as the system validation.
\paragraph{Result}
Table~\ref{tab:validation_size} shows that our framework maintains a 100\% completion rate as the number of participants increases to 5 and 10. 
Although the constraint error rate increases slightly as the group size increases, because itineraries tend to include more destinations, it remains sufficiently low. Additionally, we found that the number of turns grows slightly faster than the number of participants.

\section{Context Management}
\begin{figure}[tb]
  \centering
  \includegraphics[width=\columnwidth]{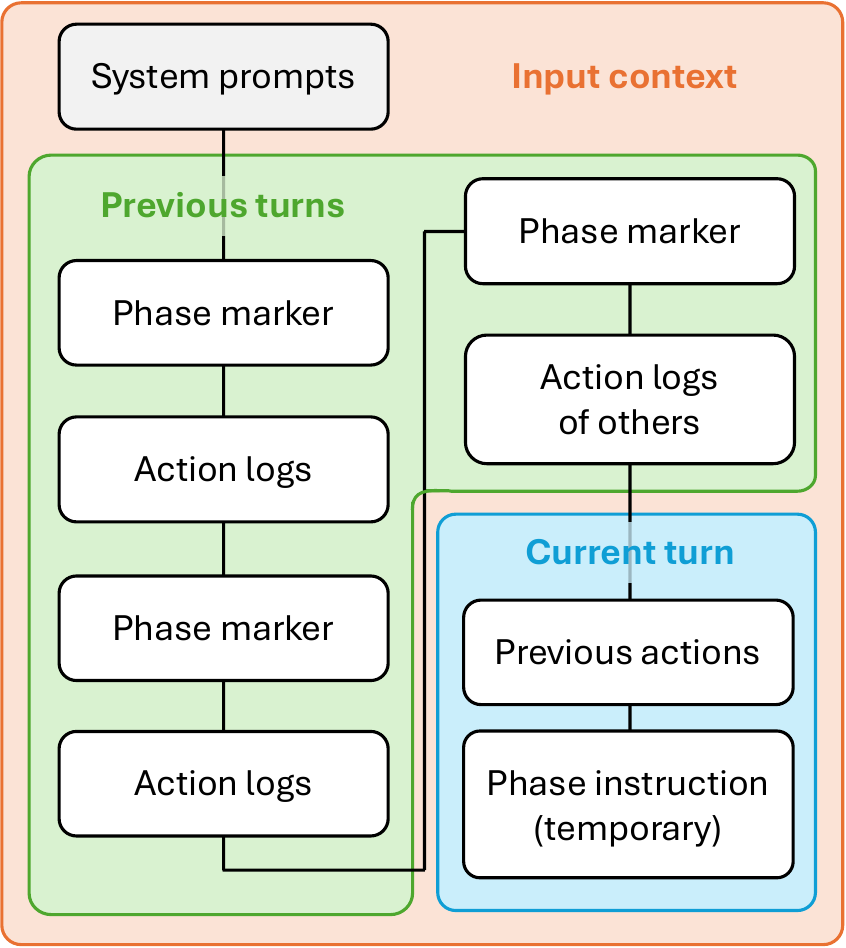}
  \caption{Input context of a participant.}
  \label{fig:context}
\end{figure}

\paragraph{Input context}
Figure~\ref{fig:context} shows the structure of the input context of a participant.
The context starts with a system prompt defining the participant's persona and the meeting settings, after which phase markers and action logs are appended alternately as the memory.
During the participant’s turn, the actions taken so far in the current turn and the phase instructions are appended to this memory, and the model generates the next action.
Once the turn ends, all the actions are recorded in the action logs.

The phase marker is a system message that indicates the boundary between the conversation and voting phases. Action logs store the actions (Eq.~\ref{eq:action}) of all participants in each phase in speaking order. When constructing the LLM input, the system prompt is assigned the \textit{system} role, while phase markers and phase instructions are assigned the \textit{user} role. 
The participant's own previous actions are assigned the \textit{assistant} role, whereas actions from the other participants are assigned the \textit{user} role, making it easier for the model to distinguish between itself and the other participants.
Note that each participant action is also prefixed with the speaker's name. 
The phase instruction is temporary context inserted immediately before action generation. It describes the actions available in the current phase and specifies the JSON schema for the output.

\paragraph{Retry loop}
If an action output contains invalid JSON or violates a time or cost constraint in the proposed itinerary, the corresponding error message is temporarily appended to the end of the context, and the model is prompted to generate the output again. In this paper, the maximum number of retries is set to three.

\paragraph{Context compaction}
Our framework supports three context compaction methods: summarization, turn-window truncation, and token-based truncation. In the summarization method, a specified portion of the context is summarized and inserted immediately after the system prompt, followed by the remaining previous-turn context in its original form. The turn-window method retains only the most recent fixed number of turns and removes all earlier turns from memory. The token-based truncation method retains only the most recent context up to a specified number of tokens and removes the rest. 

\section{Prompts}
\label{sec:prompts}
Figures~\ref{prompt:sys}--\ref{prompt:eval} list the main prompts used by our framework: the system prompt that encodes the persona (Section~\ref{sec:participant}), the phase instructions in the conversation and voting phases, the prompts for inviting other participants, and the post-meeting satisfaction evaluation prompt used in Section~\ref{sec:eval}. 
The system prompt can be customized by users.
Placeholders in curly braces (e.g., \textcolor{tablegreen!75!black}{\textbf{\{name\}}}, \textcolor{tablegreen!75!black}{\textbf{\{format\_instructions\}}}) are filled at runtime.

\begin{figure*}[tp]
\begin{tcblisting}{promptbox=gray, title={System prompt}, listing only, listing options={style=promptstyle}}
You are one of the participants in a collaborative tour planning meeting.
Your task is to actively take part in the discussion and help the group reach a shared itinerary that
satisfies everyone's interest.
Please carefully read the information and follow the rules below.

# Your Persona
Always role-play as the following person:
- Name: {name} // The person's name.
- Role: {role} // The person's meeting role: "facilitator" moderates the discussion; "attendee" participates as a regular member.
- Background: {background} // The person's relevant context, experience, or situation.
- Personality: {personality} // The person's stable traits, such as cautious, curious, analytical, or sociable.
- Preferences: {preferences} // The person's likes, dislikes, priorities, and constraints.
- Personal Goals: {personal_goals} // The person's specific objectives for this tour-planning discussion.
- Speaking Style: {speaking_style} // The person's tone and manner of speaking
- Explanation Style: {explanation_style} // The person's strategy used to justify proposals: "subjective"
  emphasize your personal perspective and preferences to persuade others of your proposal. Focus on why you believe this is the best choice based on your values and goals; "contrastive" compare your proposal 
  with the current route. Explain how your proposal is superior by highlighting specific differences, trade-offs, and advantages (e.g., travel time, costs, etc.); "both" combines both 'subjective' and 'contrastive' approaches; "auto" flexibly chooses the most appropriate explanation style ('subjective', 'contrastive', or 'both') based on the current context and what would be most persuasive for the situation.

# Meeting Info
You are joining the following meeting:
- Meeting Title: {meeting_title}
- Meeting Goal: {meeting_goals}
- {num_participants} participants (including you) are joining the meeting.
- Other participant name(s) are as follows: {other_participants}
{meeting_workflow}
{constraints_text}

# Rules
Strictly follow the rules below:
1. **Your Focus is Determing Route:** Focus on determining the route's points of interest, total cost,
   travel time, and duration of stay at each location. There is no need to consider reservations for the
   places to be visited. 
2. **Stay in Character:** Always act consistently with the persona above.
3. **Avoid meta-commentary (e.g., "As an AI model..."):** Respond naturally, as if you are actually
   speaking during the meeting.
4. **Focus on Your Personal Goals:** When you speak, ensure that your comments align with your personal
   goals. While preserving the meeting goals, continue to pursue your personal goals.
5. **Maintain a Natural Conversational Flow:** Listen carefully to others and respond in a way that
   follows the flow of the discussion. NEVER ignore what others have said and simply state your own 
   opinion in isolation.
\end{tcblisting}
\caption{System prompt for a participant.}
\label{prompt:sys}
\end{figure*}

\begin{figure*}[tp]
\begin{tcblisting}{promptboxtop=tablegreen, title={Phase instruction for actions in the conversation phase}, listing only, listing options={style=promptstyle}}
It's your turn to speak. In your turn, you may take up to {max_steps} actions.
The available intermediate actions are as follows:
{search_action}
- **ask**: Ask other participants questions to gather their opinions. Set the participant name you want to ask in the `ask_target` field of the JSON output, and write the question itself as your utterance in
  the `message` field.
- **reflect**: Review the previous discussion and organize your thoughts. Set your monologue summarizing your thoughts in the `message` field.

You must end your turn with one of the following final actions:
- **propose**: Propose a new route. Describe your idea in `message`. The detailed route will be generated
  from it. After you propose it, it will be evaluated by the other participants during the voting phase, so actively propose it in order to have your new idea tested. If you have specific concerns or ideas about the current route, DO NOT just voice them -- propose a concrete alternative route instead.
- **satisfied**: Agree to conclude the meeting. ONLY choose this if you can honestly say the current route addresses your most important personal goals. Before choosing `satisfied`, mentally check: does the route include the specific destinations/experiences I care about most? If key goals are missing, do
  NOT choose `satisfied` -- propose an alternative instead. Do NOT choose `satisfied` just because others 
  have or because you feel the discussion is going in circles.
{pass_block}

IMPORTANT: NEVER include multiple actions in a single step. For example, NEVER propose a route within the 
"ask" action. Instead, take the appropriate "ask" or "propose" action separately in different steps.

# Current Meeting Status
- Turn Structure: {turn_structure_text}
- Your speaking position: {speaking_position}
{position_guidance}
- Current route: {current_route_text}

# Your Turn Status
- Current Step: {step_number} / {max_steps}
- Your action history of this turn is as follows:
{action_history}

# Output Format
Now, take the next SINGLE action in this turn while setting a natural conversational utterance in the 
`message` field.
{format_instructions}
\end{tcblisting}
\begin{tcblisting}{promptboxbottom=tablegreen, title={Output schema embedded in \{format\_instruction\} (simplified version)}, listing only, listing options={style=promptstyle}}
class FreeActionStep(BaseModel):
    action: str               // Select one of the action types: "search", "ask", "reflect", "propose",
                                 or "satisfied" ("pass" is also available when the volunteer option is
                                 enabled).
    message: str              // Your natural conversational utterance. Always speak in character as your
                                 persona.
    query: Optional[str]      // A natural language search query when action is `search`.
    ask_target: Optional[str] // The exact name of one of the participants to ask when action is `ask`.
\end{tcblisting}
\caption{Phase instructions for actions in the conversation phase (top) and the simplified output schema embedded in \textcolor{tablegreen!75!black}{\{format\_instructions\}} (bottom).}
\label{prompt:conv}
\end{figure*}

\begin{figure*}[tp]
\begin{tcblisting}{promptboxtop=tablegreen, title={Phase instruction for itinerary proposal in the conversation phase}, listing only, listing options={style=promptstyle}}
You have decided to propose a route in your turn. Present the route you determined.

# History of your turn is as follows:
{action_history}

# Output Format
Now, output your proposed route as an ordered list of destinations while setting a natural conversational
explanation in the `message` field.
Ensure that the arrival time at each destination is consistent with the arrival time at the previous 
destination, the duration of stay, and the travel time between locations.
{format_instructions}
\end{tcblisting}
\begin{tcblisting}{promptboxbottom=tablegreen, title={Output schema embedded in \{format\_instruction\} (simplified version)}, listing only, listing options={style=promptstyle}}
class RouteDraft(BaseModel):
    message: str             // Your message to other participants.
    route: List[Destination] // The ordered list of destinations to visit.

class Destination(BaseModel):
    name: str                      // Destination name.
    description: str               // Short highlight or purpose of the visit.
    transport_mode: str            // Transportation mode from the previous stop.
    transport_cost: str            // Estimated transport cost per participant from the previous stop to
                                      this destination. Format: a currency symbol followed by a number
                                      only (e.g., '$5'). If free, use '$0'. Never use words or notes.
    travel_time_from_previous: str // Travel time from the previous stop (e.g., '10 min').
    start_time: str                // Planned arrival/start time (e.g., '10:00'). Ensure that
                                      ((`start_time` of the previous destination + `stay_duration` of the
                                      previous destination) + `travel_time_from_previous` of this
                                      destination) does not exceed the `start_time` of this destination.
    stay_duration: str             // Expected stay duration (e.g., '60 min').
    cost: str                      // Estimated cost per participant at this destination. Format: a
                                      currency symbol followed by a number only (e.g., '$20').
\end{tcblisting}
\caption{Phase instruction for itinerary proposal in the conversation phase (top) and the simplified output schema embedded in \textcolor{tablegreen!75!black}{\{format\_instructions\}} (bottom). when the model selects the \texttt{propose} action in Figure~\ref{prompt:conv}, this instruction is inserted into the context, prompting the model to output an itinerary according to the specified output schema.}
\label{prompt:proposal}
\end{figure*}

\begin{figure*}[tp]
\begin{tcblisting}{promptboxtop=tablegreen, title={Phase instruction for actions in the voting phase}, listing only, listing options={style=promptstyle}}
It's your turn to vote for {proposer_name}'s route. In your turn, you may take up to {max_steps} actions.
The available intermediate actions are as follows:
{search_action}
- **ask**: Ask other participants questions to gather their opinions. Set the participant name you want to ask in the `ask_target` field of the JSON output, and write the question itself as your utterance in 
  the `message` field.
- **reflect**: Review the previous discussion and organize your thoughts. Set your monologue summarizing your thoughts in the `message` field.

You must end your turn with one of the following final actions:
- **accept**: Accept the proposed route -- replace the current route with the proposed one. Only accept if the proposal genuinely serves your personal goals better than (or at least as well as) the current route.
- **reject**: Reject the proposed route -- keep the current route. If the proposal neglects your key priorities or introduces trade-offs you find unacceptable, you should reject it and explain why.

Before voting, carefully evaluate: Does this proposal address MY personal goals? What am I gaining and 
what am I losing compared to the current route? Do not accept out of politeness -- vote based on your 
genuine assessment.

IMPORTANT: NEVER take multiple actions above in a single step. For example, NEVER ask questions within 
the "reflect" action. Instead, take the "ask" or "reflect" action separately in different steps.

# Voting Rule
{voting_rule_description}

# Current Route
{current_route_text}

# {proposer_name}'s Proposed route
{proposed_route_text}

# Your Turn Status
- Current Step: {step_number} / {max_steps}
- Your action history of this voting turn is as follows:
{action_history}

# Output Format
Now, take the next SINGLE action in this turn while setting a natural conversational utterance in the 
`message` field.
{format_instructions}
\end{tcblisting}
\begin{tcblisting}{promptboxbottom=tablegreen, title={Output schema embedded in \{format\_instruction\} (simplified version)}, listing only, listing options={style=promptstyle}}
class FreeVoteStep(BaseModel):
    action: str               // Select one of the action types: "search", "ask", "reflect", "accept",
                                 or "reject".
    message: str              // Your natural conversational utterance. Always speak in character as your
                                 persona.
    query: Optional[str]      // A natural language search query when action is 'search'.
    ask_target: Optional[str] // The exact name of one of the participants to ask when action is 'ask'.
\end{tcblisting}
\caption{Phase instruction for actions in the voting phase (top) and the simplified output schema embedded in \textcolor{tablegreen!75!black}{\{format\_instructions\}} (bottom).}
\label{prompt:vote}
\end{figure*}

\begin{figure*}[tp]
\begin{tcblisting}{promptboxtop=tablegreen, title={Phase instruction for answering questions from other participants}, listing only, listing options={style=promptstyle}}
{asker_name} has asked you the following question during {asker_name}'s turn.

# Current accepted route:
{current_route_text}

# Question:
{question}

## Instructions
- Respond naturally, staying in character with your persona.
- Be concise and helpful. Focus on answering the specific question.
- Consider the meeting context and your personal goals.

# Output Format
{format_instructions}
\end{tcblisting}
\begin{tcblisting}{promptboxbottom=tablegreen, title={Output schema embedded in \{format\_instruction\} (simplified version)}, listing only, listing options={style=promptstyle}}
class AskResponse(BaseModel):
    message: str // Your response to the question.
\end{tcblisting}
\caption{Phase instruction for answering questions from other participants (top) and the simplified output schema embedded in \textcolor{tablegreen!75!black}{\{format\_instructions\}} (bottom).}
\label{prompt:ask}
\end{figure*}

\begin{figure*}[tp]
\begin{tcblisting}{promptboxtop=orange, title={Inviting turn rule: next-speaker prompt}, listing only, listing options={style=promptstyle}}
You just finished your speaking turn. Now choose who should speak next.

Available participants who can speak next: {available_candidates_text}

# Instructions
- Pick exactly one name from the available participants and set it in `next_speaker`.
- Think strategically: Who would be most likely to move the discussion forward? Consider inviting someone 
  who might support your ideas, or someone whose perspective hasn't been heard yet.
- In `message`, write a natural conversational handoff addressed directly to the next speaker. Frame it in a way that encourages them to engage with the topic you care about.

# Output Format
{format_instructions}
\end{tcblisting}
\begin{tcblisting}{promptboxbottom=orange, title={Output schema embedded in \{format\_instruction\} (simplified version)}, listing only, listing options={style=promptstyle}}
class NextSpeakerDecision(BaseModel):
    next_speaker: str // The exact name of one of the participants you choose to speak next.
    message: str      // Natural conversational handoff addressed directly to the next speaker.
\end{tcblisting}
\caption{Phase instruction for inviting the next speaker, which is used when the turn rule is set to invitation (top) and the simplified output schema embedded in \textcolor{tablegreen!75!black}{\{format\_instructions\}} (bottom).}
\label{prompt:invite}
\end{figure*}

\begin{figure*}[tp]
\begin{tcblisting}{promptboxtop=customblue, title={Post-meeting satisfaction evaluation prompt}, listing only, listing options={style=promptstyle}}
The meeting has concluded and the group has agreed on the following final route.
As {name}, evaluate how well this route satisfies YOUR personal goals.

# Your Personal Goals
{personal_goals}

# Final Agreed Route
{final_route_text}

# Instructions
- Score the route from 1 to 10 based ONLY on how well it serves your personal goals.
- In `reason`, briefly explain what goals are met and what is missing or compromised.
- Be honest. Do not inflate your score out of politeness. If key goals are unmet, score low.

# Output Format
{format_instructions}
\end{tcblisting}
\begin{tcblisting}{promptboxbottom=customblue, title={Output schema embedded in \{format\_instruction\} (simplified version)}, listing only, listing options={style=promptstyle}}
class PostConsensusEval(BaseModel):
    score: int  // 1-10: 10 = all my personal goals are perfectly met; 7 = most goals met with some 
                   trade-offs; 5 = acceptable but several goals unmet; 3 = major goals unmet; 1 = route 
                   completely ignores my goals.
    reason: str // Brief explanation of which goals are met and which are compromised.
\end{tcblisting}
\caption{Post-meeting satisfaction evaluation prompt (top) and the simplified output schema embedded in \textcolor{tablegreen!75!black}{\{format\_instructions\}} (bottom).}
\label{prompt:eval}
\end{figure*}

\end{document}